\documentclass[letterpaper]{article} 
\usepackage{aaai21}  
\usepackage{times}  
\usepackage{helvet} 
\usepackage{courier}  
\usepackage[hyphens]{url}  
\usepackage{graphicx} 
\usepackage{natbib}  
\usepackage{caption} 
\usepackage{amsmath,amssymb,amsfonts,bm,urwchancal,tikz,standalone,booktabs,subfig}
\usepackage[mathscr]{euscript}
\usepackage[symbols,record]{glossaries-extra}

\graphicspath{{img/}}

\usetikzlibrary{arrows.meta}

\GlsXtrLoadResources[src={symbols,acronyms},selection=all]

\makeindex

\begin{document}

\title{Specializing Communication in Heterogeneous Deep Multi-Agent Reinforcement Learning using Agent Class Information}
\author {
	Douglas De Rizzo Meneghetti\qquad{}Reinaldo Augusto da Costa Bianchi\\
}
\affiliations {
	FEI University Center \\
	Humberto de Alencar Castelo Branco Ave., 3972-B\\
	São Bernardo do Campo, SP, Brazil, 09850-901\\
	douglasrizzo@fei.edu.br, rbianchi@fei.edu.br
}

\maketitle
\begin{abstract}
	\begin{quote}
		Inspired by recent advances in agent communication with graph neural networks, this work proposes the representation of multi-agent communication capabilities as a directed labeled heterogeneous agent graph, in which node labels denote agent classes and edge labels, the communication type between two classes of agents. We also introduce a neural network architecture that specializes communication in fully cooperative heterogeneous multi-agent tasks by learning individual transformations to the exchanged messages between each pair of agent classes. By also employing encoding and action selection modules with parameter sharing for environments with heterogeneous agents, we demonstrate comparable or superior performance in environments where a larger number of agent classes operates.
	\end{quote}
\end{abstract}

\section{Introduction}

In partially observable multi-agent settings, communication among agents may help mitigate the uncertainty with relation to which state the agent currently is, given its observation. Communication has been achieved in deep multi-agent reinforcement learning by multiple means: with agents directly sharing their observations~\cite{Sunehag2018} or learning message passing mechanisms via backpropagation. This was initially achieved with multi-layer perceptrons~\cite{Sukhbaatar2016}, bidirectional recurrent neural networks~\cite{Peng2017b}, considering communication as part of agent actions~\cite{Foerster2016c} or allowing an agent to learn which agents to communicate with in the environment, without restraints~\cite{Hoshen2017,Jiang2018,Malysheva2019,Das2019}.

A recent spike in interest in the areas of graph neural networks and geometric deep learning resulted in works which explore the use of operations previously targeted towards supervised and semi-supervised learning on graphs to achieve agent communication~\cite{Wang2018,Agarwal2019a,Malysheva2019,Jiang2020}. However, most of these works focus in tasks with homogeneous agents, that is, agents that have the same observation and action spaces and follow the same policy.

Many tasks present agents that would either benefit from learning specific policies or that have completely different observations and actions. An example of the former is the RoboCup robot soccer leagues~\cite{Chalup2019}, in which teams of homogeneous robots may specialize into offensive, defensive and goalkeeping roles. Examples of the later include mixed robot teams, such as aerial drones used for mapping and terrestrial robots tasked with navigation, as well as real-time strategy (\gls{rts}) games. In an \gls{rts} game, each unit belongs to an unit type, which dictates its skills, strengths and weaknesses and, consequently, its policy.

Another property of the aforementioned examples is the fact that a collection of heterogeneous agents can be group into classes (robot soccer roles, such as offense, defense and goalie, or game unit types), where agents from the same class may be considered homogeneous among themselves.

Given these scenarios, this work proposes a neural network model that learns specialized communication protocols between classes of agents. The method first creates a directed labeled agent communication graph, in which node labels represent agent classes and edge labels represent a communication channel between agents of two classes. The neural network model then employs relational graph convolutions~\cite{Schlichtkrull2018} as a message passing mechanism between agents, before estimating \gls{Q} values or all actions. We also show how heterogeneous classes of homogeneous agents can share parameters in parts of the model, accelerating learning.

\section{Research Background}

When focusing on fully-cooperative, partially observable multi-agent tasks, one way to formalize the environments is through decentralized partially observable Markov decision processes \glspl{decpomdp}~\cite{Amato2015}. A \gls{decpomdp} is defined as a tuple \(<\gls{agents}, \gls{S}, \gls{actions}, \gls{O}, \gls{transition}, \gls{reward_function}, O>\), where

\begin{itemize}
	\item \gls{agents}: a finite set of agents,
	\item \gls{S}: a finite set of states,
	\item \(\gls{actions}_u\): finite set of actions for agent \(u\),
	\item \(\gls{O}_u\): a finite set of observations for agent \(u\),
	\item \gls{transition}: a transition probability function \(\gls{transition}(s'|s, \bm{a})\) mapping the joint actions chosen by all agents in state \(s\) to the probability of transitioning to state \(s'\),
	\item \gls{reward_function}: a shared reward function \(\gls{reward_function}(s,\bm{a})\),
	\item \(O\): an observation model \(O(\bm{o}|s',\bm{a})\) which dictates the probabilities that agents will observe \(\bm{o}\) in \(s'\) after taking \(\bm{a}\).
\end{itemize}

Given a pair of agents \(u_1\) and \(u_2\), they are considered homogeneous when \(\gls{actions}_{u_1} = \gls{actions}_{u_2}\), \(\gls{O}_{u_1} = \gls{O}_{u_2}\) and they can share the same policy without affecting outcomes. If at least one of these conditions is not true, then they are considered to be heterogeneous.

\subsection{Deep Multi-Agent Reinforcement Learning}

Recent advances in the area of deep multi-agent reinforcement learning that contributed to this work are the network architectures trained with off-policy algorithms to control multiple agents in fully cooperative, partially observable environments. Reinforced Inter-Agent Learning and Differentiable Inter-Agent Learning~\cite{Foerster2016c} model agents as deep recurrent \gls{Q}-Networks (DRQN)~\cite{Hausknecht2015} to address partial observability, while mixing techniques such as value decomposition networks (\gls{vdn})~\cite{Sunehag2018} and QMIX~\cite{Rashid2018} train a multi-agent neural network by combining agents' \gls{Q} values under the assumptions of additivity and monotonicity, respectively, to achieve superior performance in fully cooperative tasks. Agents then choose actions according to their individual \gls{Q} values, achieving what is called centralized training and decentralized execution.

Especially in the case of \gls{vdn}, the joint value function of a team of cooperative agents is given by \[\gls{Q}_{\gls{agents}}(\bm{o}, \bm{a}) = \sum_{u \in \gls{agents}} \gls{Q}_u(o_u, a_u).\]

\subsection{Graph Neural Networks}

Graph neural networks (\gls{gnn}) ~\cite{Gori2005,Scarselli2009} are models specialized in processing input data structured as graphs or manifolds. \glspl{gnn} are composed of multiple types of operations~\cite{Zhou2018}. However, one class of operations over graphs that is particularly interesting in the MARL setting are graph convolutions~\cite{Defferrard2017,Kipf2017}. These operations can be interpreted as message passing mechanisms between nodes of a graph~\cite{Gilmer2017a} and, while their main applications have been in supervised and semi-supervised learning in graph data sets, they have also found use in reinforcement learning tasks, as message passing mechanisms between agents~\cite{Sukhbaatar2016,Malysheva2019,Jiang2020}.

In the family of graph convolution operations, one that is particularly interesting in this work are relational graph convolutions (\gls{rgcn})~\cite{Schlichtkrull2018}. \glspl{rgcn} operate over graphs defined as \(\gls{graph} = (\gls{vertices}, \gls{edges}, \gls{relations})\), where \gls{vertices} represents a set of nodes containing individual feature vectors, \gls{edges} is a set of edges and \gls{relations} a set of possible relations between nodes.

The feature vector of node \(i\) in layer \(l+1\) is given by
\begin{equation}\label{eq:rgcn}
	\vec{v}_i^{\,(l+1)} = \sigma \left( \sum_{r \in \gls{relations}} \sum_{j \in \gls{neighbors}^{r}_{i}} \frac{1}{c_{i,\,r}} \bm{W}^{(l)}_r \vec{v}^{\,(l)}_j + \bm{W}^{(l)}_0 \vec{v}^{\,(l)}_i \right),
\end{equation}
where \(\sigma\) is a nonlinear activation function, \(r\) is the index of the relation between nodes \(i\) and \(j\), \(\gls{neighbors}^{r}_{i}\) are the neighbors of \(i\) under connected by edges with relation index \(r\) and \(\bm{W}^{(l)}_r\) is a parameter matrix specific to relation \(r\) in layer \(l\).

\section{Related Work}

Other works have already modeled agent communication capability as graphs. \citet{Agarwal2019a} use a Transformer attention mechanism~\cite{Vaswani2017} to let agents learn the relevance of neighbors' messages, while DGN~\cite{Jiang2020} concatenate the outputs of multiple graph convolution layers to form agents' observations, under the rationale that each subsequent layer captures information from further away in the graph.

MAGNet~\cite{Malysheva2019} uses a neural network to generate weighted edges in the agent graph. This graph generation network is pre-trained in environments which possess default agents and its output is then used by the multi-agent reinforcement learning network to learn policies for all agents.

While the previous mentioned works have only dealt with homogeneous agents, i.e. agents that work with the same \gls{actions} and \gls{O} sets as well as learn the same policy, NerveNet~\cite{Wang2018} models a creature's skeleton as a group of heterogeneous agents by categorizing joints with equal functionality as agents of the same type. A static graph of the creature's skeleton is then generated and messages are passed among nodes by using specialized MLPs for each pair of node types.

In previous work~\cite{Meneghetti2020b}, a neural network architecture was presented to deal with graphs composed of heterogeneous agents and entities, in which node vectors represented an entity's features instead of its observations. Due to partial observability of the environment, the graph structure proposed in that work was not capable of representing all information necessary for agents to learn well-performing policies, an issue which has been addressed in this work.

The current work is different from \citet{Agarwal2019a} and DGN by taking into account partial observations, using LSTM layers in the agent networks and training agents using an episodic replay buffer~\cite{Hausknecht2015} instead of a buffer containing transitions~\cite{Mnih2015}. This work also differs from MAGNet and other works in which an agent may freely choose to communicate with any other agent in the environment~\cite{Foerster2016c,Hoshen2017,Jiang2018,Das2019}, as well as NerveNet, in which the agent communication graph is fixed, given the environment. In our work, communication capabilities are dictated by the environment and may change between states, being out of the control of the agents.

Lastly, it is worth noting that the use of agent classes to model MDPs has previously been proposed in object-oriented MDPs~\cite{Wasser2008} and its multi-agent variant~\cite{daSilva2019}.

\section{Representing states as heterogeneous agent graphs}

In this section, we introduce the concept of a heterogeneous agent communication graph \(\gls{graph} = (\gls{vertices}, \gls{edges},\gls{C}. \gls{relations})\). Each node in the set of nodes \gls{vertices} represents an agent in the set of agents \gls{agents}, so, for the rest of this work, individual agents and nodes will both be represented by \(u\).

A directed arc \((u_1, u_2) \in \gls{edges}\) denotes the capability of \(u_1\) to communicate with \(u_2\). By making \gls{graph} a directed graph, we are able to represent one-way communication between agents when necessary.

In some tasks, the team of agents may have full communication among themselves, making \gls{graph} a complete graph. In other tasks, an agent may be limited to communicate only with agents that are geographically close or with which some kind of communication channel is available. In this work, we make no assumptions regarding agent communication capabilities and assume they may change from state to state.


The set \gls{C} represents a collection of agent classes. All agents \(u \in \gls{agents}\) belong to a single class in \gls{C}, in such a way that, while \gls{agents} may be considered a set of heterogeneous agents, the subset of all agents belonging to a class \(c \in \gls{C}\), denoted by \(\gls{agents}_c\), is homogeneous.

Let \(\gls{C}(u)\) be a function that returns the class of \(u\). An arc \((u_1, u_2)\) is labeled \((\gls{C}(u_1), \gls{C}(u_2))\), indicating a communication channel from an agent of class \(\gls{C}(u_1)\) to an agent of class \(\gls{C}(u_2)\). This, coupled with the fact that agent class information is encoded in \gls{graph} through the use of node labels, makes \gls{graph} a labeled graph.

The set of all possible inter-agent communication channel types is denoted by \gls{relations} and is defined as all possible ordered pairs of agent classes \((c_1, c_2), c1 \in C, c2 \in C\), totaling \(|C|^2\) possible types.

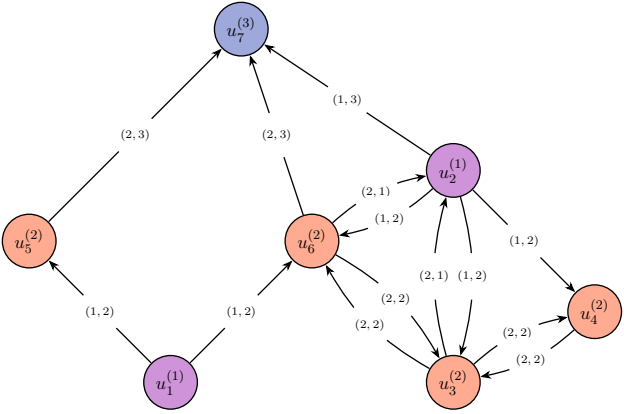
\begin{figure}
	\centering
	\caption{A heterogeneous agent graph, represented as a directed labeled graph}
	\label{fig:heterogeneous-agent-graph}
	\definecolor{c1}{RGB}{206,147,216}
\definecolor{c2}{RGB}{255,171,145}
\definecolor{c3}{RGB}{159,168,218}
\begin{tikzpicture}
	\begin{scope}[every node/.style={fill=c1,circle,thick,draw,font=\large}]
		\node (u1) at (3, 3)      {\(u_1^{(1)}\)};
		\node (u2) at (9, 7.5)    {\(u_2^{(1)}\)};
	\end{scope}
	\begin{scope}[every node/.style={fill=c2,circle,thick,draw,font=\large}]
		\node (u3) at (9, 3)      {\(u_3^{(2)}\)};
		\node (u4) at (12, 4.5)   {\(u_4^{(2)}\)};
		\node (u5) at (0, 6)      {\(u_5^{(2)}\)};
		\node (u6) at (6, 6)      {\(u_6^{(2)}\)};
	\end{scope}
	\begin{scope}[every node/.style={fill=c3,circle,thick,draw,font=\large}]
		\node (u7) at (4.5, 10.5) {\(u_7^{(3)}\)};
	\end{scope}

	\begin{scope}[>={Stealth[black]},
		every node/.style={fill=white,circle,font=\scriptsize},
		every edge/.style={draw=black,thick}]
		\path [->] (u3) edge [bend left=15] node {\((2,2)\)} (u4);
    \path [->] (u4) edge [bend left=15] node {\((2,2)\)} (u3);
    
		\path [->] (u6) edge [bend left=15] node {\((2,1)\)} (u2);
		\path [->] (u2) edge [bend left=15] node {\((1,2)\)} (u6);
    
    \path [->] (u2) edge                node {\((1,3)\)} (u7);
    
    \path [->] (u6) edge                node {\((2,3)\)} (u7);
    
    \path [->] (u5) edge                node {\((2,3)\)} (u7);
    
    \path [->] (u3) edge [bend left=15] node {\((2,2)\)} (u6);
		\path [->] (u6) edge [bend left=15] node {\((2,2)\)} (u3);
    
    \path [->] (u1) edge node {\((1,2)\)} (u6);
    
    \path [->] (u1) edge                node {\((1,2)\)} (u5);
    
    \path [->] (u3) edge [bend left=15] node {\((2,1)\)} (u2);
		\path [->] (u2) edge [bend left=15] node {\((1,2)\)} (u3);
    
    \path [->] (u2) edge                node {\((1,2)\)} (u4);
	\end{scope}
\end{tikzpicture}
\end{figure}

Figure~\ref{fig:heterogeneous-agent-graph} presents an example of such a graph. In it, node colors and superscripts represent an agent's class and bidirectional communication is represented by a pair of arcs.

\section{Heterogeneous Communication in Deep Multi-Agent Reinforcement Learning}

\begin{figure*}
	\centering
	\caption{Neural network architecture for the heterogeneous agent setting.}
	\label{fig:hmagnet}
	\includegraphics[width=\textwidth]{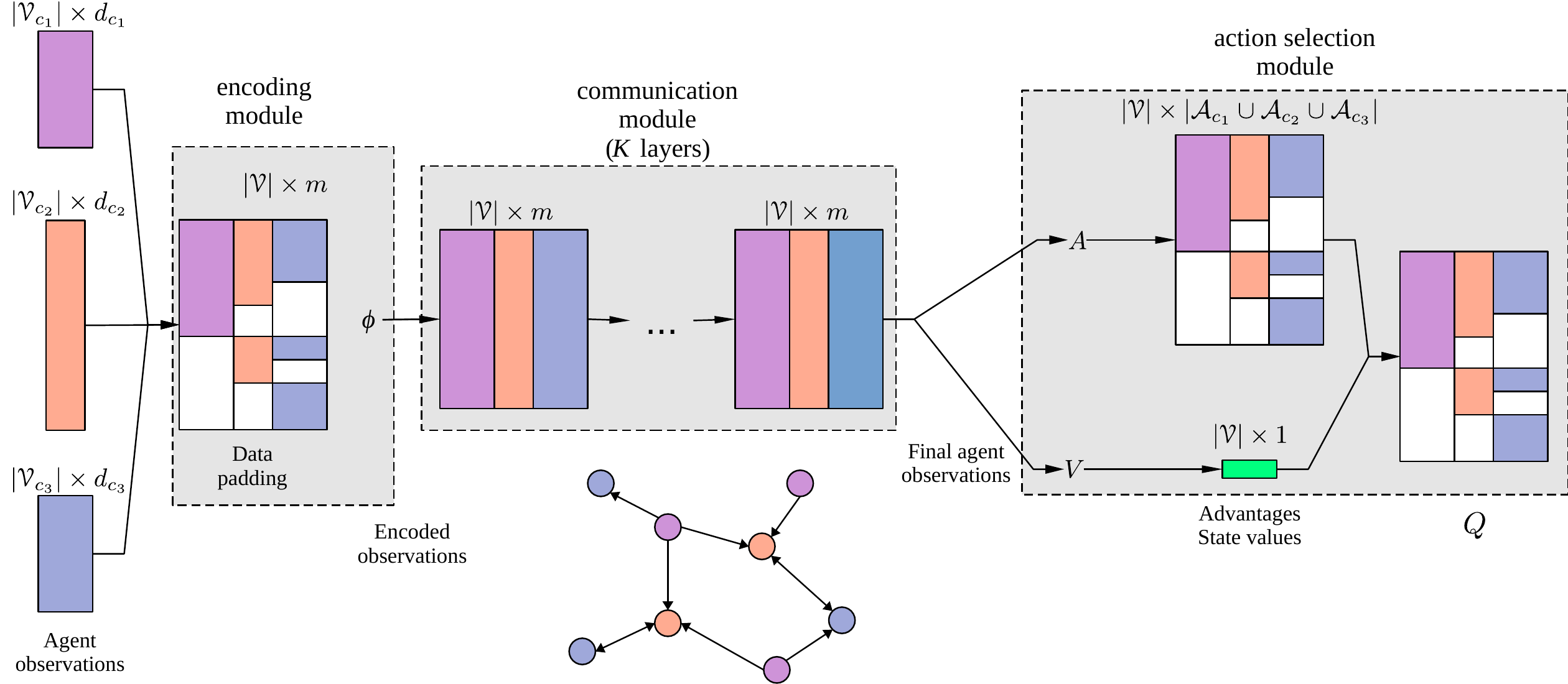}
\end{figure*}

Figure~\ref{fig:hmagnet} presents the proposed neural network architecture. It is composed of three modules. The input data is composed of \(\bm{o}\), the observations of all agents. The encoding module applies a function \gls{phi} to \(\bm{o}\), normalizing its dimensions and enhancing the expressiveness of the model.

In the communication module, the agent graph structure is used for message passing among agents through the use of \gls{rgcn} layers. Equation \eqref{eq:rgcn} shows how communication between different pairs of classes of agents is learned and performed differently. By representing each agent-class pair \((c_1, c_2), c1 \in C, c2 \in C\) as a relation \(r \in \gls{relations}\), an \gls{rgcn} layer is able to model message passing between each pair of agent class using individual parameter matrices, allowing the overall network to shape the way agents of specific classes communicate with each other independently.

The resulting vectors after applying \(K\) graph convolution layers are taken as the combination of each agent's observation and the information received by its neighbors. They are then used as input to the action selection module, which also employs a single function that approximates \gls{Q} values for all agents, sharing parameters in an analogous way to the encoding function \gls{phi}.

\subsection{Parameter sharing for teams of heterogeneous agents}

In \citet{Meneghetti2020b}, both the encoding and action selection modules were composed of multiple functions \(\gls{phi}_c\) and \(\gls{Q}_c\),\( c \in \gls{C}\), specialized in agent classes with observation vectors of different dimensions, as well as action sets of different sizes. However, we found that using the same input and output dimensions for all agent classes and employing zero padding in the unused variables allowed for the use of a single function in each case, achieving parameter sharing and faster training of the model.

Furthermore, when using the same action selection network for multiple heterogeneous agents, we have to account for the fact that the network will estimate \gls{Q} values for the joint set of actions of all agent classes. When choosing an action according to their policies, each agent must ignore values for actions outside of its action set. These \gls{Q} values must also be ignored when calculating the temporal difference error that trains the network.

Let \gls{actions_c} denote the action set of agent class \(c\) and \(\gls{actions} = \bigcup_{c \in \gls{C}} \gls{actions_c}\) the joint set of agent actions. Whenever the network estimates \gls{Q} for an action \(a \in \gls{actions} - \gls{actions}_{C(u)}\) for an agent \(u\), we ignore it during action selection and consider its contribution to the TD error as 0.

This parameter sharing approach for heterogeneous agents was formalized in~\citet{Terry2020} for the general case in which all agents in a team are heterogeneous. Here we apply the same rationale when a heterogeneous group of agents can be grouped into classes of homogeneous agents.

\section{Experiments}

The proposed network architecture was tested in the \acrlong{smac} (\gls{smac}) environment~\cite{Samvelyan2019}. A collection of scenarios was chosen to evaluate the performance of the model under different circumstances, such as a scenario with a single agent class (3m), scenarios with increasing number of agent classes (3s5z and 1c3s5z) and scenarios in which both teams had the same number of units (MMM) vs an equivalent scenario in which the agent team had a disadvantage in number of units (MMM2).

Actions in the SMAC scenarios include: moving in one of four directions cardinal directions, attacking an enemy unit (select by an ID), stopping and the no-op action. The Medivac unit, present in the MMM and MMM2 scenarios, has actions to heal allied units instead of attacking enemies. Due to the parameterized nature of the attack and heal actions, units in different scenarios may have a varying number of actions.

The encoding function \gls{phi} is a single layer perceptron with 96 neurons and tanh activation; the action selection module is implemented as a Dueling Deep Recurrent \gls{Q}-Network \cite{Mnih2015,Hausknecht2015,Wang2016} with the Double \gls{Q}-Learning loss function~\cite{vanHasselt2016a}.

We tested three different variants of the communication module. The first one uses the proposed heterogeneous agent communication graph and two \gls{rgcn} layers~\cite{Schlichtkrull2018} with two bases. The second module employs two graph attention (\gls{gat}) layers~\cite{Velickovic2018} with three attention heads. \gls{gat} layers do not explicitly specialize message passing according to agent classes, but may do so implicitly if agent class information is added to the agent observation. In both cases, each layer is composed of 96 hidden units and employs the LeakyReLU activation. The last variant of the communication module is no communication at all.

To determine the orthogonality of mixing strategies (which have been applied by other works in the SMAC environment) with the proposed agent communication method, we include in our analysis one method that approximates individual value functions for each agent, known as independent \gls{Q}-Learning (\gls{iql})~\cite{Tan1993} and another one that learns the additive joint value function for the team of agents (\gls{vdn})~\cite{Sunehag2018}.

Agents use a joint \gls{epsilon}-greedy policy, in which all agents either explore with probability \gls{epsilon} or exploit with probability \(1 - \gls{epsilon}\). A linear \gls{epsilon} decay schedule is employed, whose relevant values, as well as other hyperparameters used in the experiments, are displayed in table~\ref{tbl:hyperparams}.

\begin{table}
	\centering
	\caption{Hyperparameters used in the training setting}
	\label{tbl:hyperparams}
	\begin{tabular}{rl}
		\(\hat\theta\) update interval               & \(250\) \\
		Network learning rate                        & \(2.5*10^{-4}\) \\
		L2 regularization coef.                      & \(10^{-5}\) \\
		RL discount factor \(\gamma\)                & \(0.99\) \\
		\(\epsilon_{min}\)                           & \(0.1\) \\
		\(\epsilon_{max}\)                           & \(0.95\) \\
		N. steps to finish linear \(\epsilon\) decay & 50000
	\end{tabular}
\end{table}

Every 10 thousand steps, we execute an evaluation step in which the network follows a greedy policy for 32 episodes. We log the win rate in these episodes, as well as the average number of defeated enemies (as an intermediate metric, in case agents do not win a match).

Following recommendations from the literature~\cite{Rashid2020}, we train each version of the model in each of the chosen maps five times for 1 million steps, reporting the 25, 50 and 75 percentiles of the aforementioned metrics in each combination. We also trained select versions of the model for 6 million steps on a single map (3s5z) to investigate the results of longer training periods. Each 1 million-step training procedure took from 5 to 11 eleven wall-clock hours in an Nvidia GTX 1070 GPU.

\section{Results}

\begin{figure*}
	\centering
	\caption{Win rate (left column) and number of defeated enemies (right column) for 1 million training steps (100 evaluation steps) in the tested SMAC maps}
	\label{fig:wr_de}
	\subfloat[Win rate -- 3m\label{fig:wr_3m}]{\includegraphics[width=.43\textwidth]{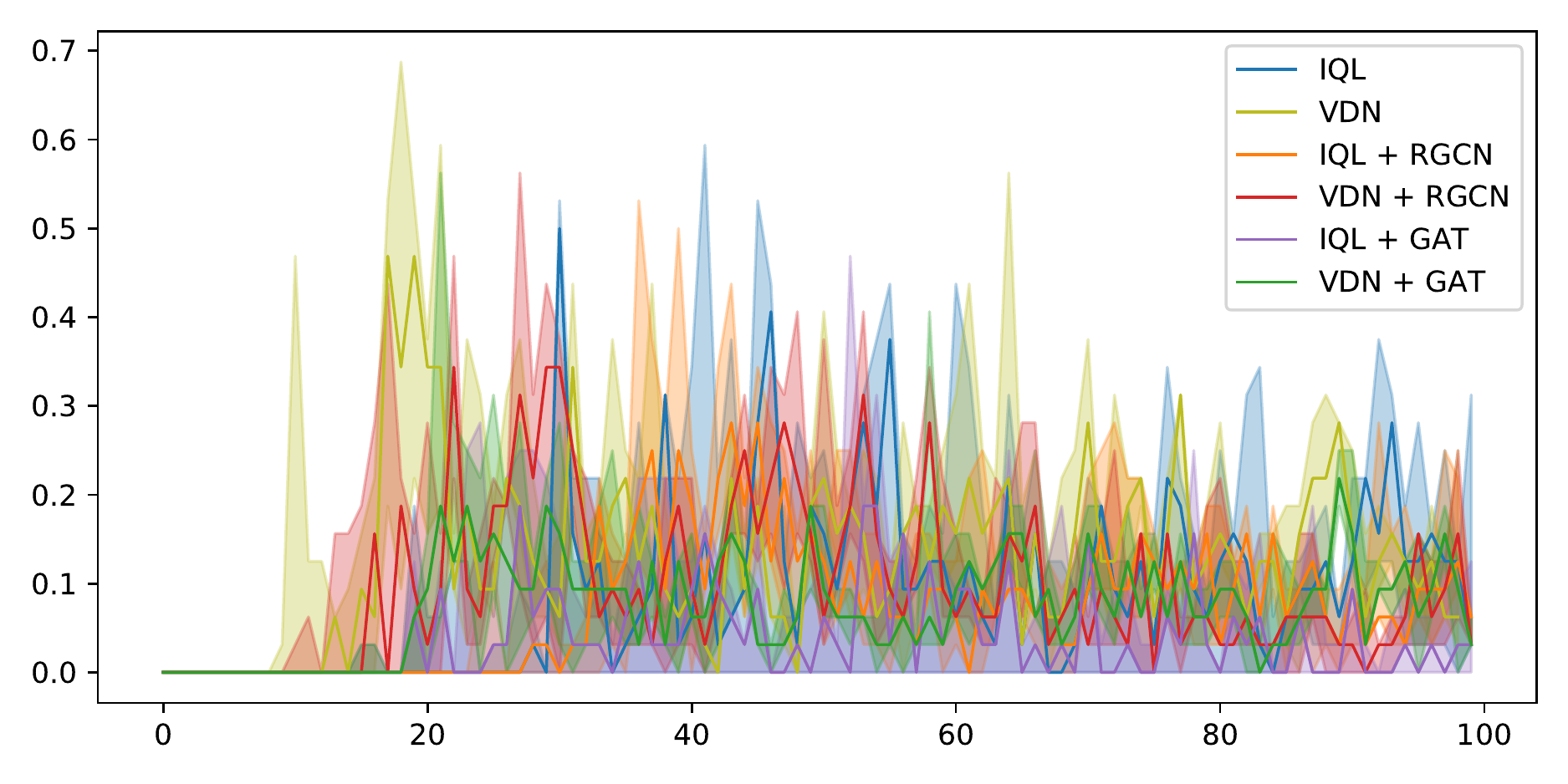}}
	\subfloat[Defeated enemies -- 3m (max 3)\label{fig:de_3m}]{\includegraphics[width=.43\textwidth]{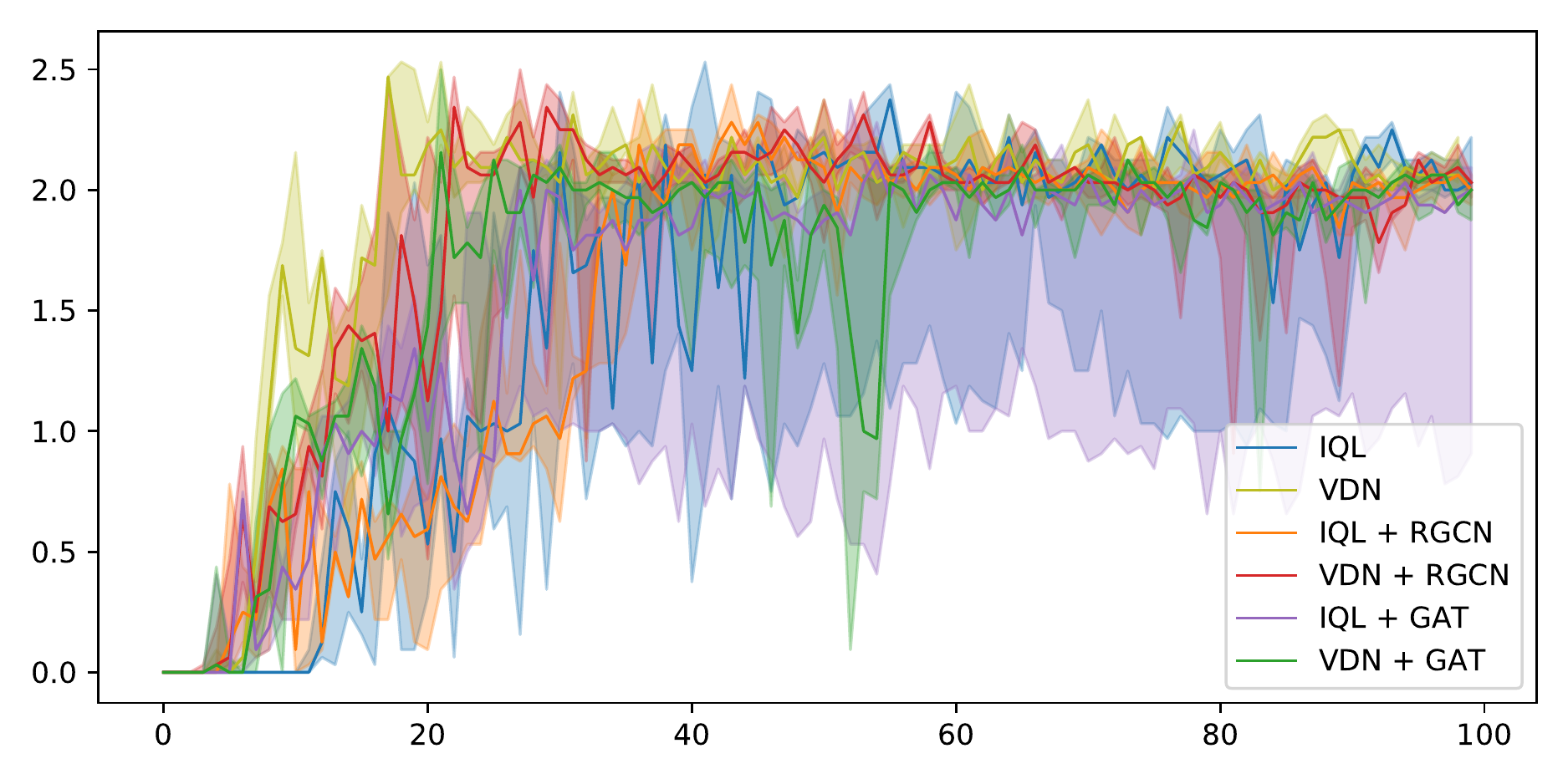}} \\
	\subfloat[Win rate -- 3s5z\label{fig:wr_3s5z}]{\includegraphics[width=.43\textwidth]{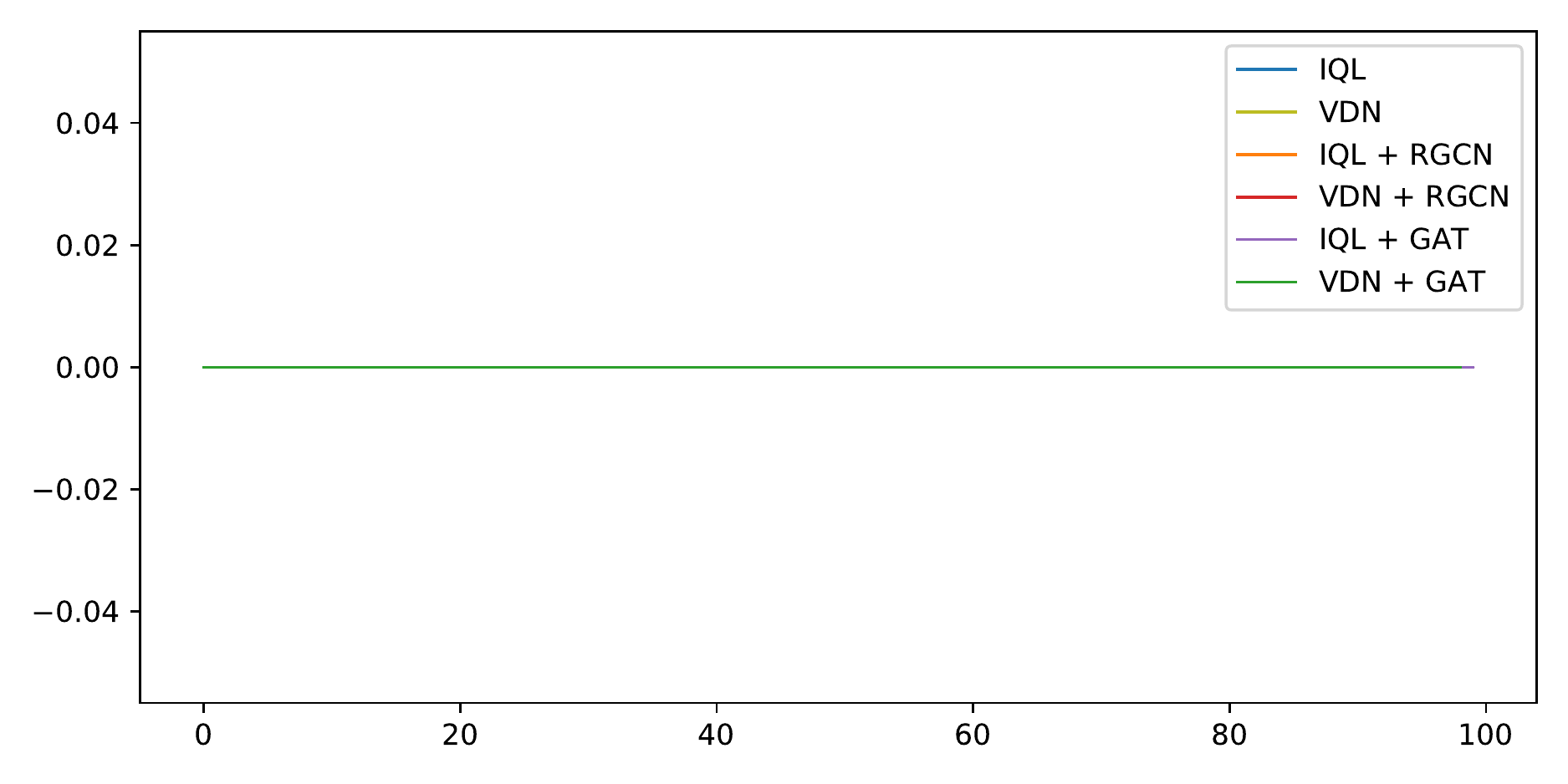}}
	\subfloat[Defeated enemies -- 3s5z (max 8)\label{fig:de_3s5z}]{\includegraphics[width=.43\textwidth]{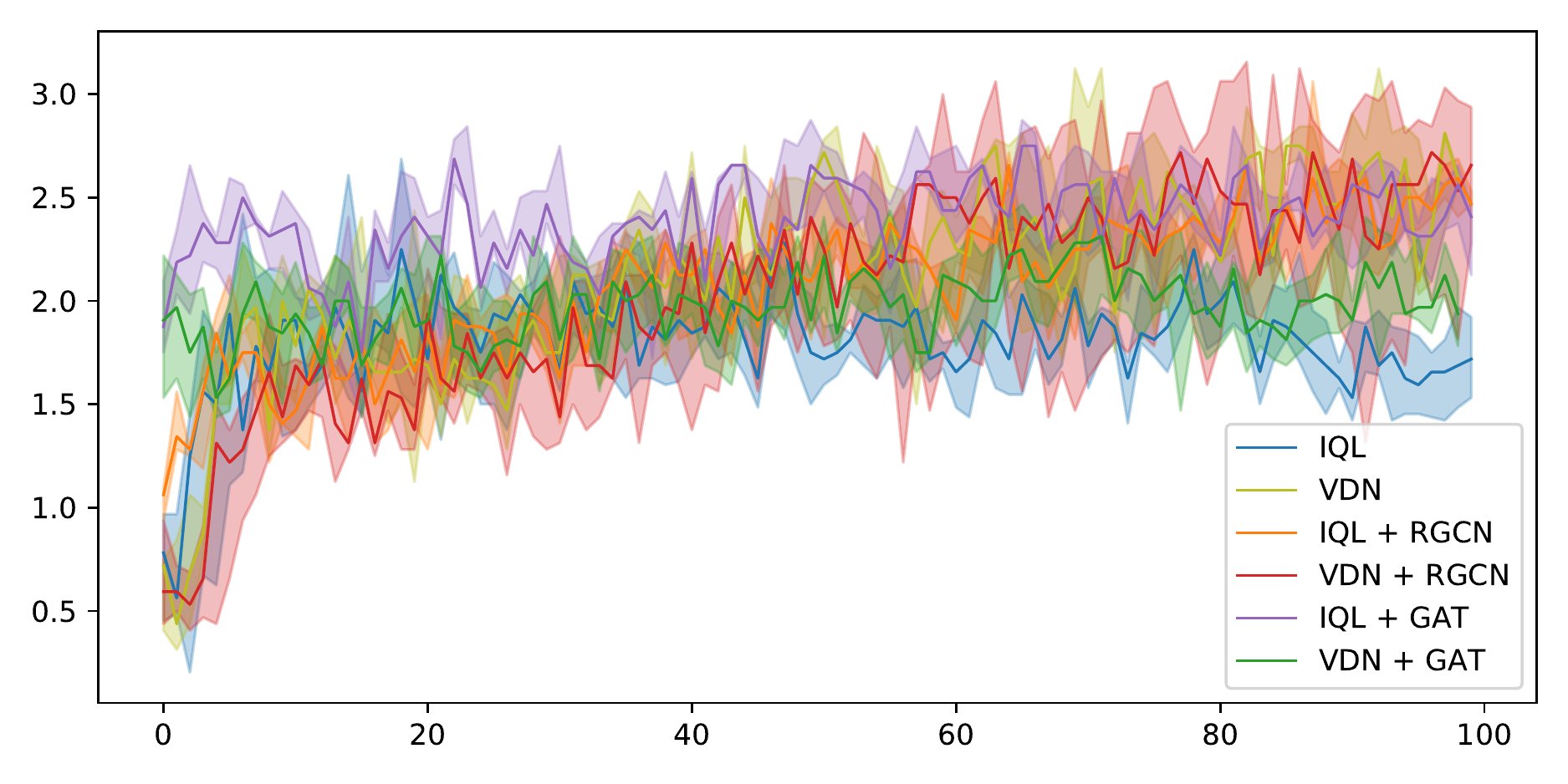}} \\
	\subfloat[Win rate -- 1c3s5z\label{fig:wr_1c3s5z}]{\includegraphics[width=.43\textwidth]{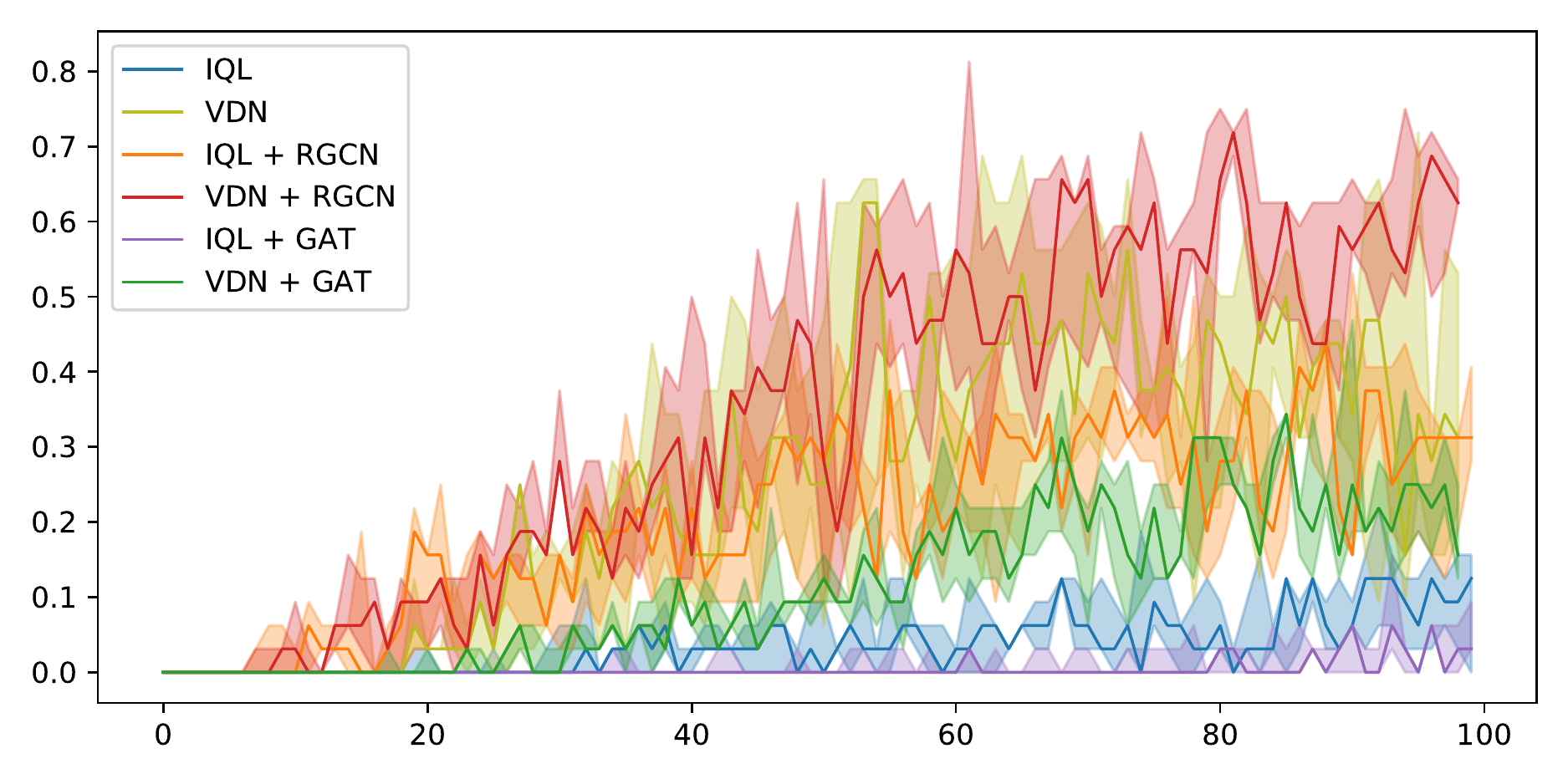}}
	\subfloat[Defeated enemies -- 1c3s5z (max 8)\label{fig:de_1c3s5z}]{\includegraphics[width=.43\textwidth]{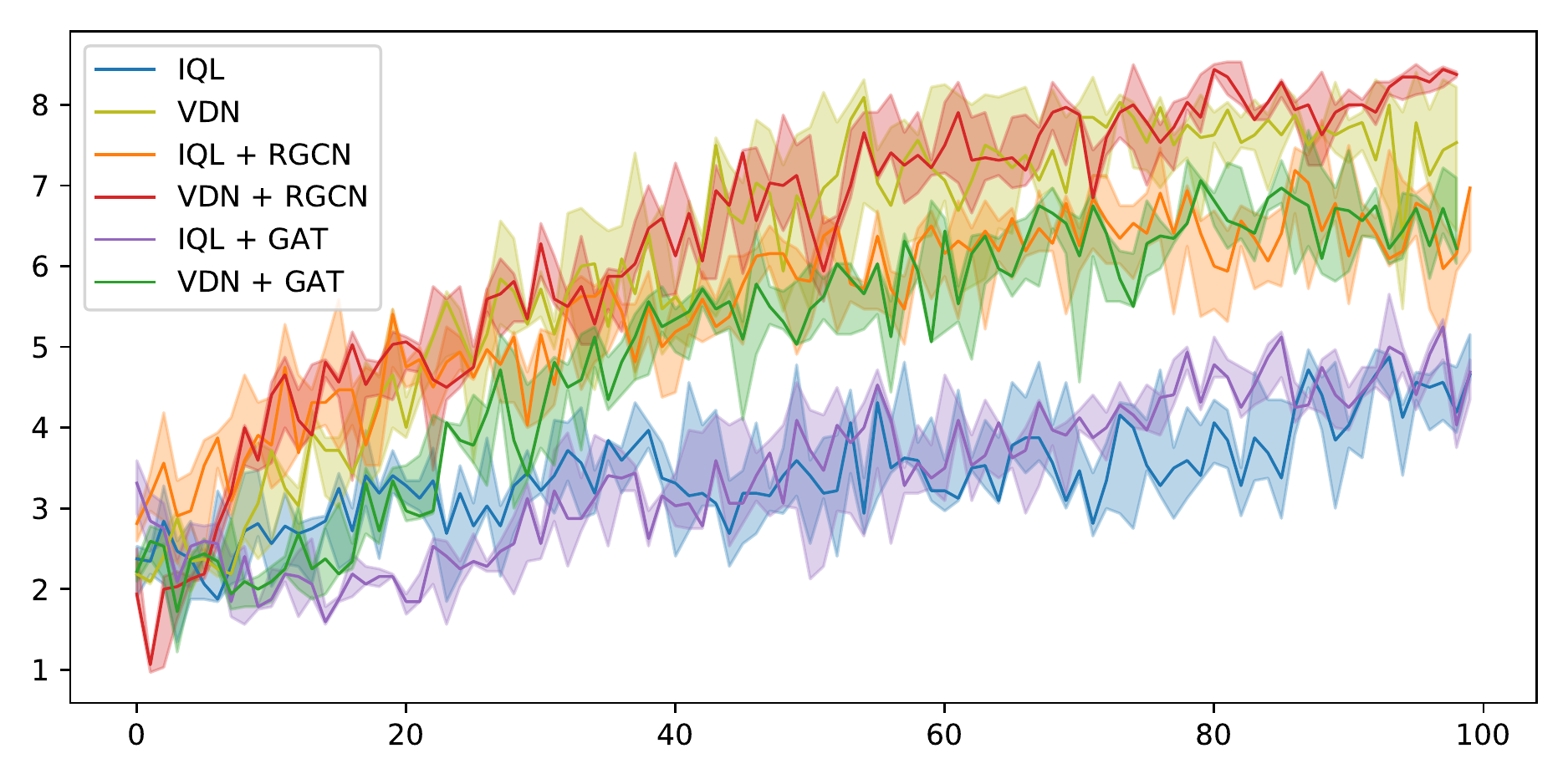}}\\
	\subfloat[Win rate -- MMM\label{wr_MMM}]{\includegraphics[width=.43\textwidth]{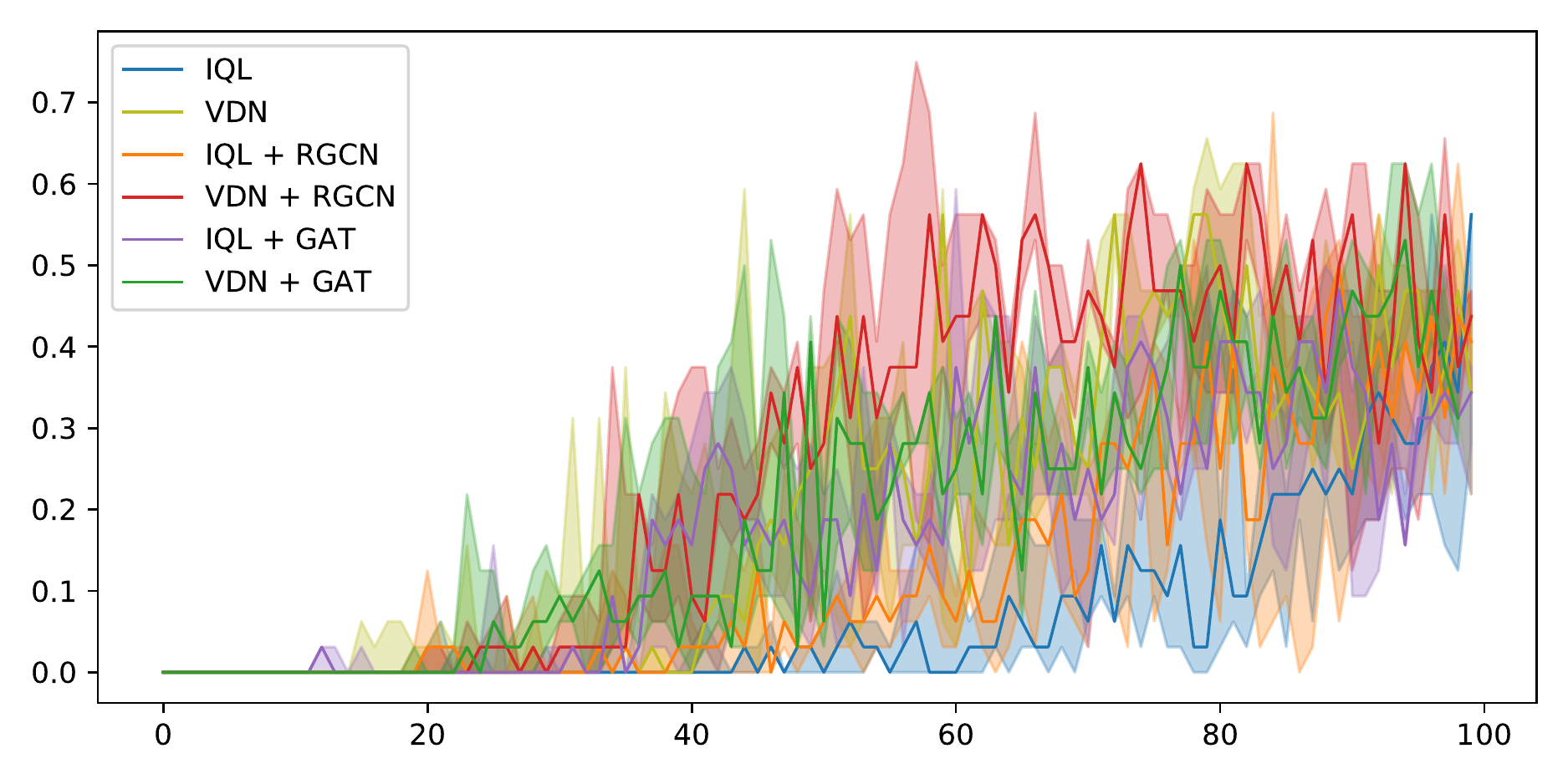}}
	\subfloat[Defeated enemies -- MMM (max 10)\label{fig:de_MMM}]{\includegraphics[width=.43\textwidth]{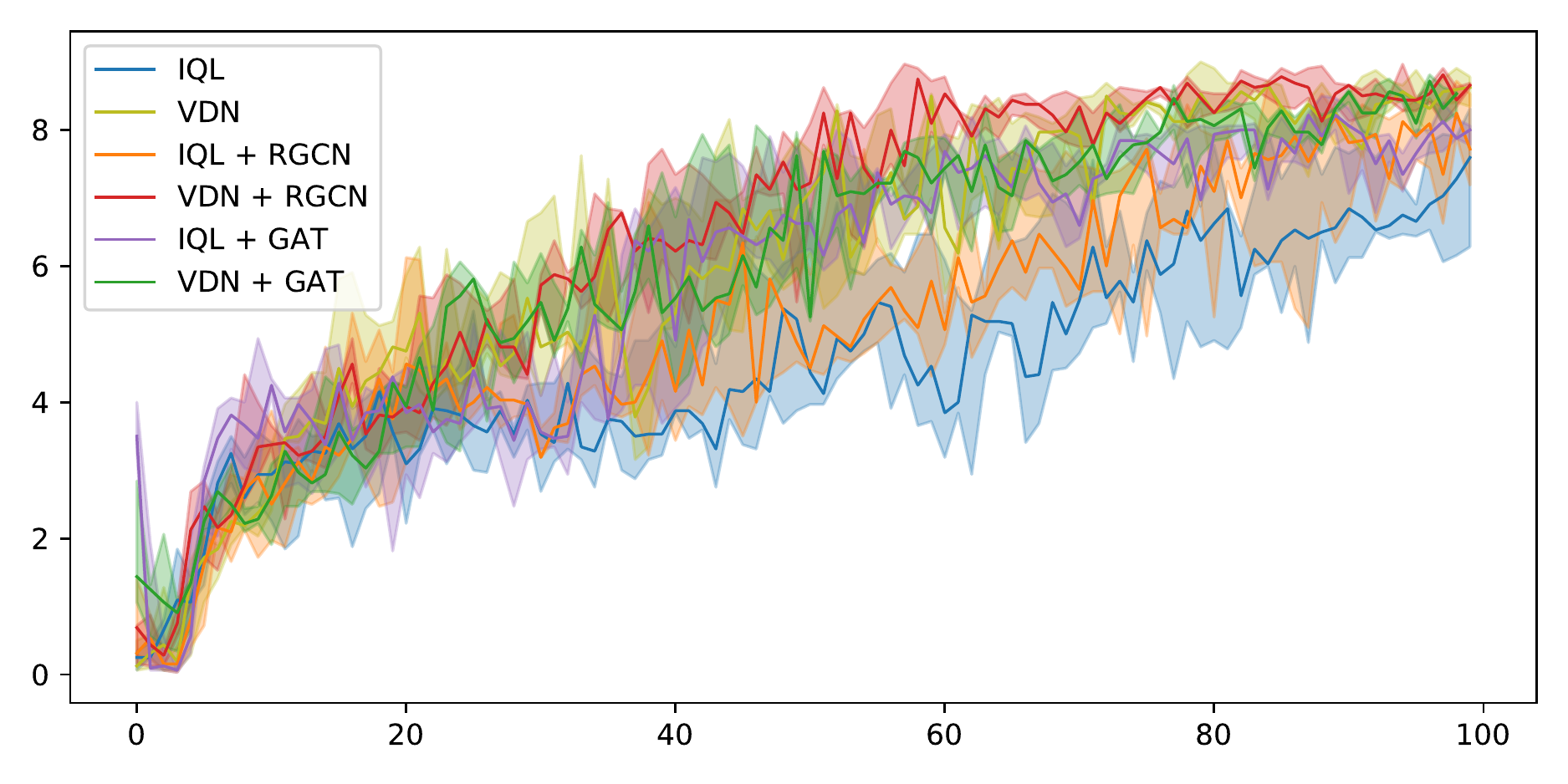}} \\
	\subfloat[Win rate -- MMM2\label{wr_MMM2}]{\includegraphics[width=.43\textwidth]{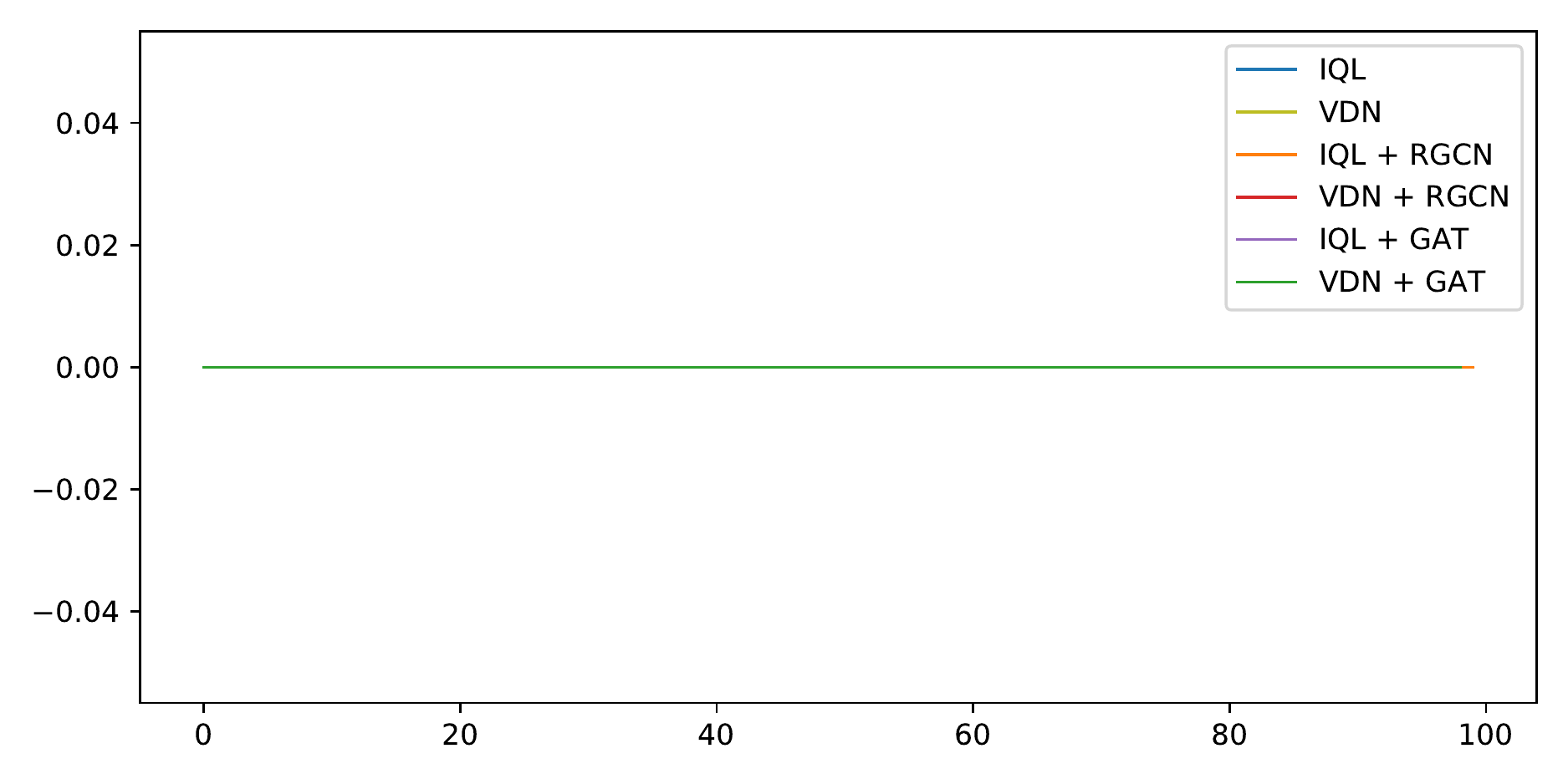}}
	\subfloat[Defeated enemies -- MMM2 (max 12)\label{fig:de_MMM2}]{\includegraphics[width=.43\textwidth]{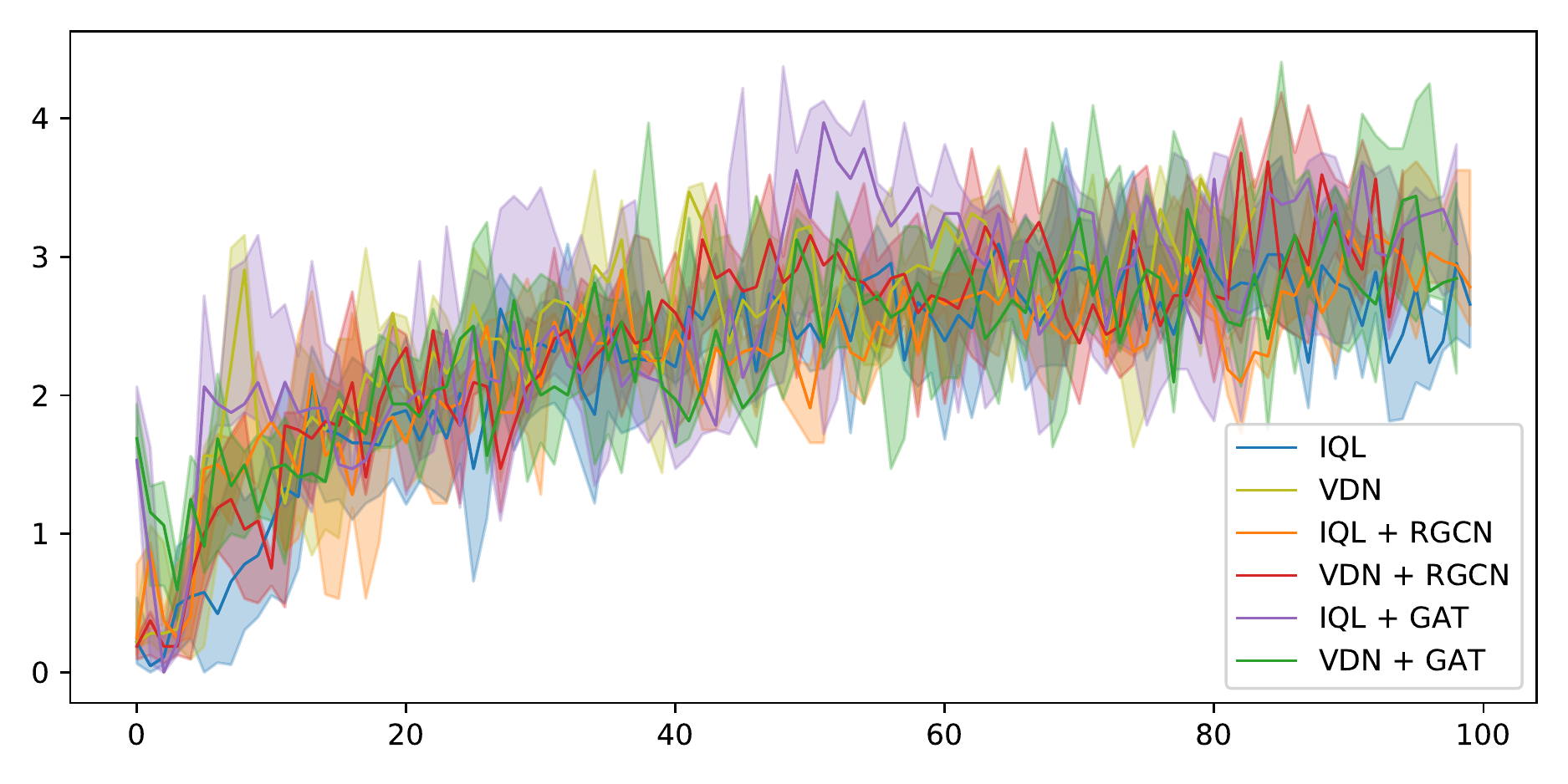}}
\end{figure*}

Figure~\ref{fig:wr_de} displays the win rate and average number of defeated enemies in the tested SMAC maps. While in other works~\cite{Rashid2018,Rashid2020}, authors present different results for equivalent algorithms (\gls{iql}, \gls{vdn} without communication) in the same maps, this may be due to the use of a different version of StarCraft.

While the number of defeated enemies in the homogeneous 3m map (\ref{fig:de_3m}) grows as training progresses, all methods seem to display equivalent performance with relation to win rate (\ref{fig:wr_3m}). The same can be said for the heterogeneous 3s5z map (which we investigate further in forthcoming results) and the MMM2 map, which is considered a hard map due to the asymmetry between the agent and enemy teams.

In the 1c3s5z which is a symmetric scenario with the most classes of agents (3), the methods display different performance. \gls{iql} with specialized communication achieved superior win rates and defeated more enemies than \gls{iql} with communication performed using an attention mechanism as well as no communication (figs. \ref{fig:wr_1c3s5z}, \ref{fig:de_1c3s5z}). In this scenario, communication using an attention mechanism seemed to harm agent performance, performing the worst of all. Lastly, we can see that the model that used a combination of specialized communication and value decomposition achieved the highest values in our metrics, implying a benefit in using specialized communication in heterogeneous multi-agent settings over no communication or attention mechanisms.

Lastly, in the MMM scenario, which also has three agent classes, although the model with specialized communication and value decomposition tended to defeat the most number of agents in average (fig.~\ref{fig:de_MMM}), all models displayed a tendency to reach equivalent final performance, implying that, while not negatively affecting performance, communication may not be necessary in all scenarios.

\begin{figure}[!htb]
	\centering
	\caption{Tukey's HSD test in different performance measures of the different methods.}
	\label{fig:tukey}
	\subfloat[Win rate -- 1c3s5z\label{fig:tukey_1}]{\includegraphics[width=.43\textwidth]{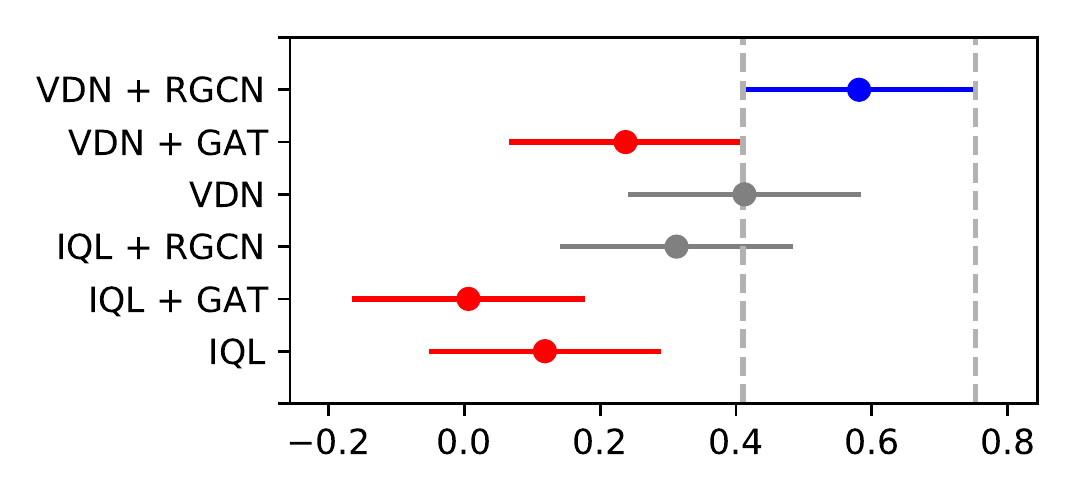}}

	\subfloat[Defeated enemies -- 3s5z\label{fig:tukey_2}]{\includegraphics[width=.43\textwidth]{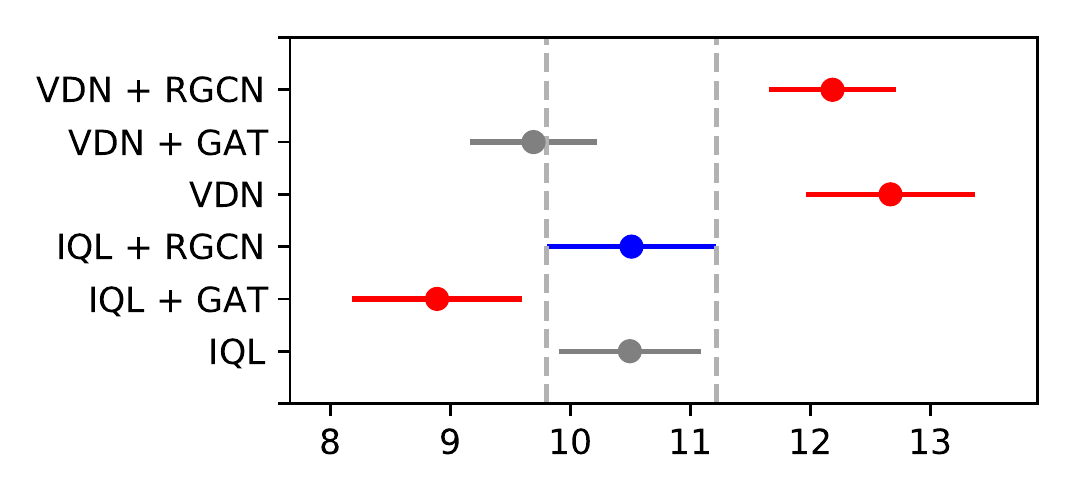}}

	\subfloat[Defeated enemies -- MMM\label{fig:tukey_3}]{\includegraphics[width=.43\textwidth]{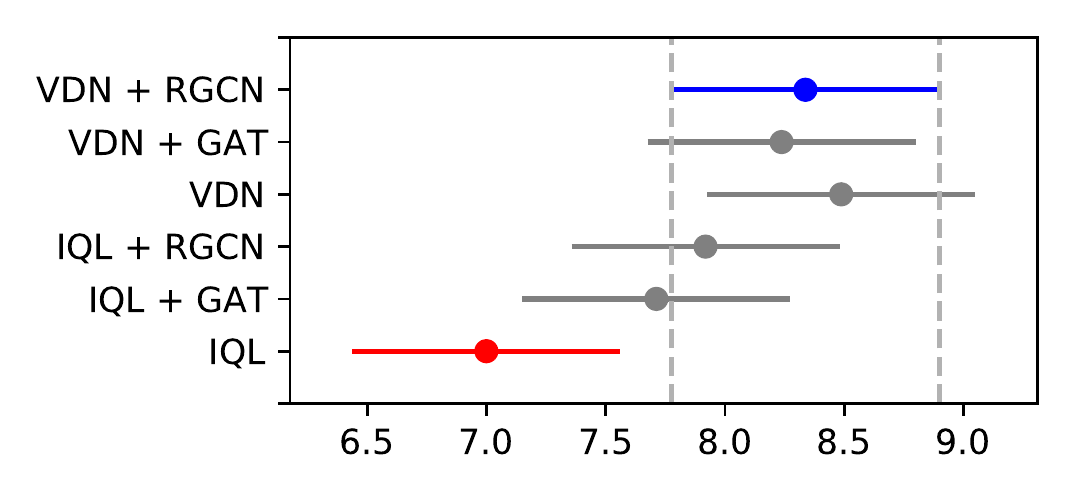}}

	\subfloat[Episode reward -- MMM2\label{fig:tukey_4}]{\includegraphics[width=.43\textwidth]{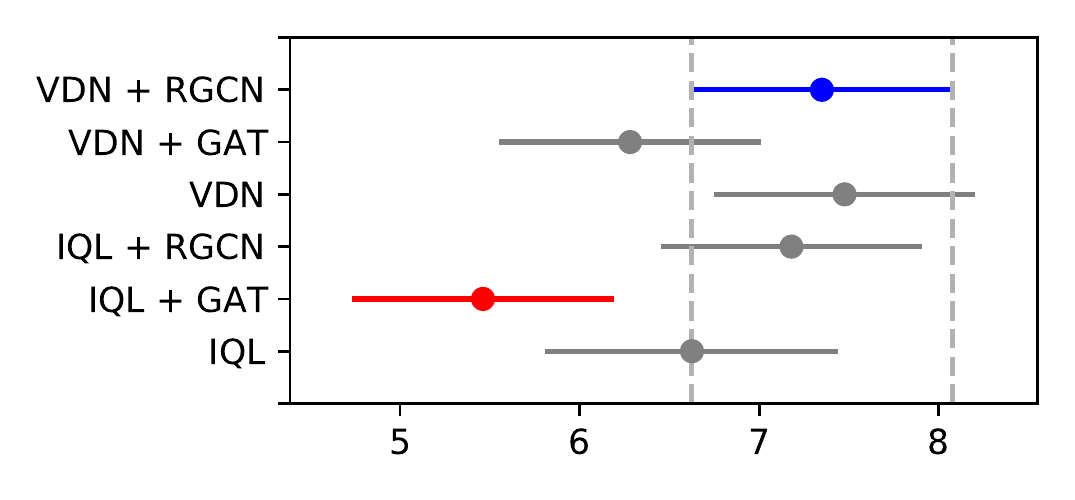}}
\end{figure}

To discover whether there are relevant differences in the performance of the methods, we employed analysis of variance on all six tested methods in each of the scenarios separately, followed by a Tukey HSD test. Figure~\ref{fig:tukey} presents results of Tukey's test applied to the data collected in the final five evaluation steps of all methods. We omit tests which are highly similar (scenarios with both a high win rate and number of defeated enemies) and tests performed in the 3m scenario, in which no statistically significant difference among the methods was observed (possibly due to agent homogeneity).

In some scenarios, there are indications that the use of specialized communication and additive mixing may be comparable (fig.~\ref{fig:tukey_1}). Other measures indicate that the use of an attention mechanism for communication may have been detrimental in some scenarios, when compared to specialized or no communication (fig.~\ref{fig:tukey_2}). Lastly, while in some scenarios it may be unclear which is the best performing method, methods that employ specialized communication are constantly in the groups of highest performance (fig.~\ref{fig:tukey_1}, \ref{fig:tukey_2}, \ref{fig:tukey_3}, \ref{fig:tukey_4}).



Given the poor win rate achieved by all models in the symmetric and heterogeneous 3s5z scenario (fig. \ref{fig:wr_3s5z}), an additional experiment was conducted, in which a selection of the models was trained for a total of 6 million steps. Models were selected in an ablative fashion, to contrast: the contribution of specialized communication over no communication; the orthogonality of the use of a mixing strategy over \gls{iql} and the performance of specialized communication over an attention mechanism.

\begin{figure}[!htb]
	\centering
	\caption{Win rate and defeated enemies for 6 million training steps (600 evaluation steps) in the \texttt{3s5z} map}
	\label{fig:6M}
	\subfloat[Win rate\label{fig:wr_6M}]{\includegraphics[width=.49\textwidth]{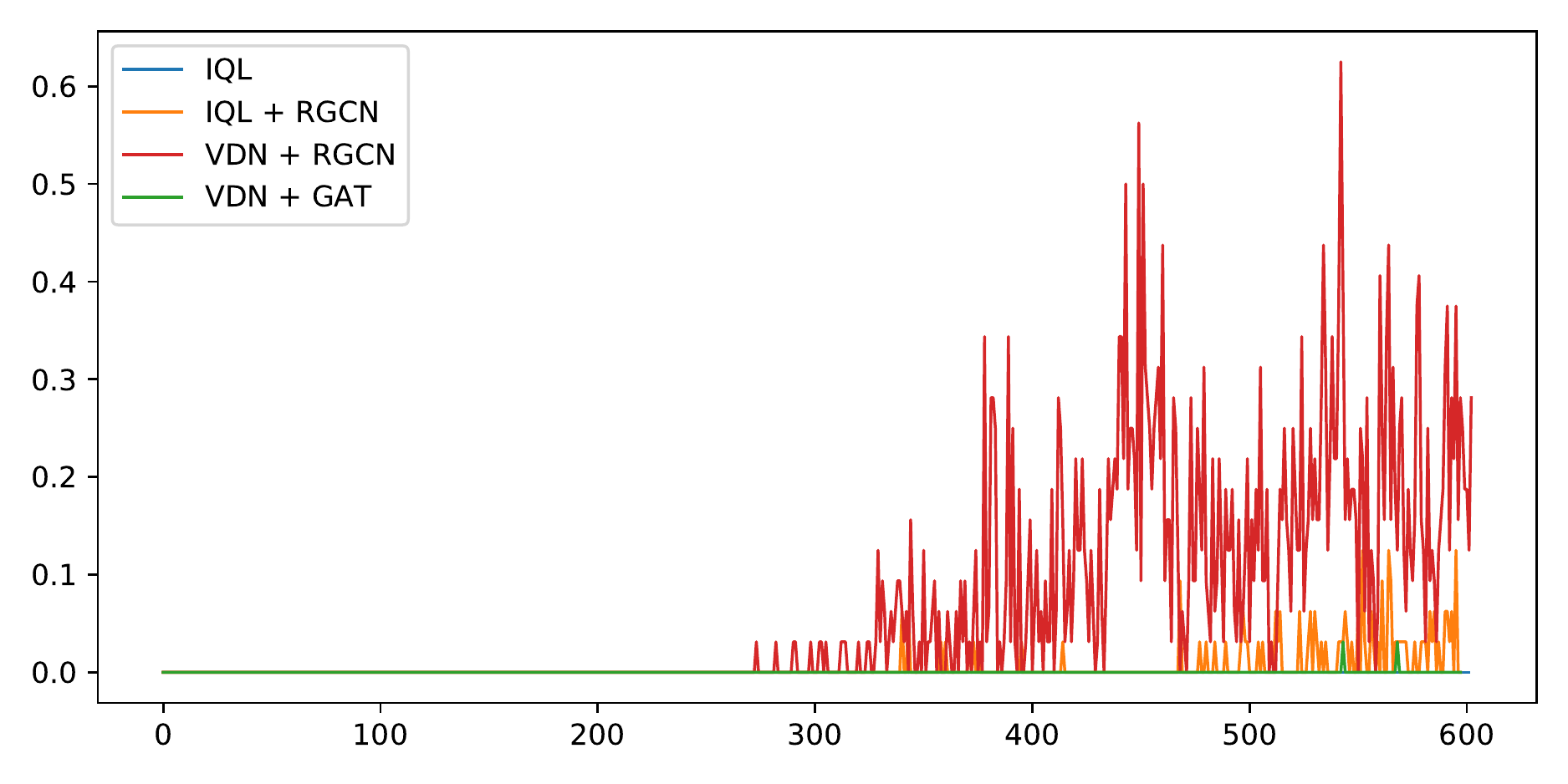}}

	\subfloat[Defeated enemies (max 8)\label{fig:de_6M}]{\includegraphics[width=.49\textwidth]{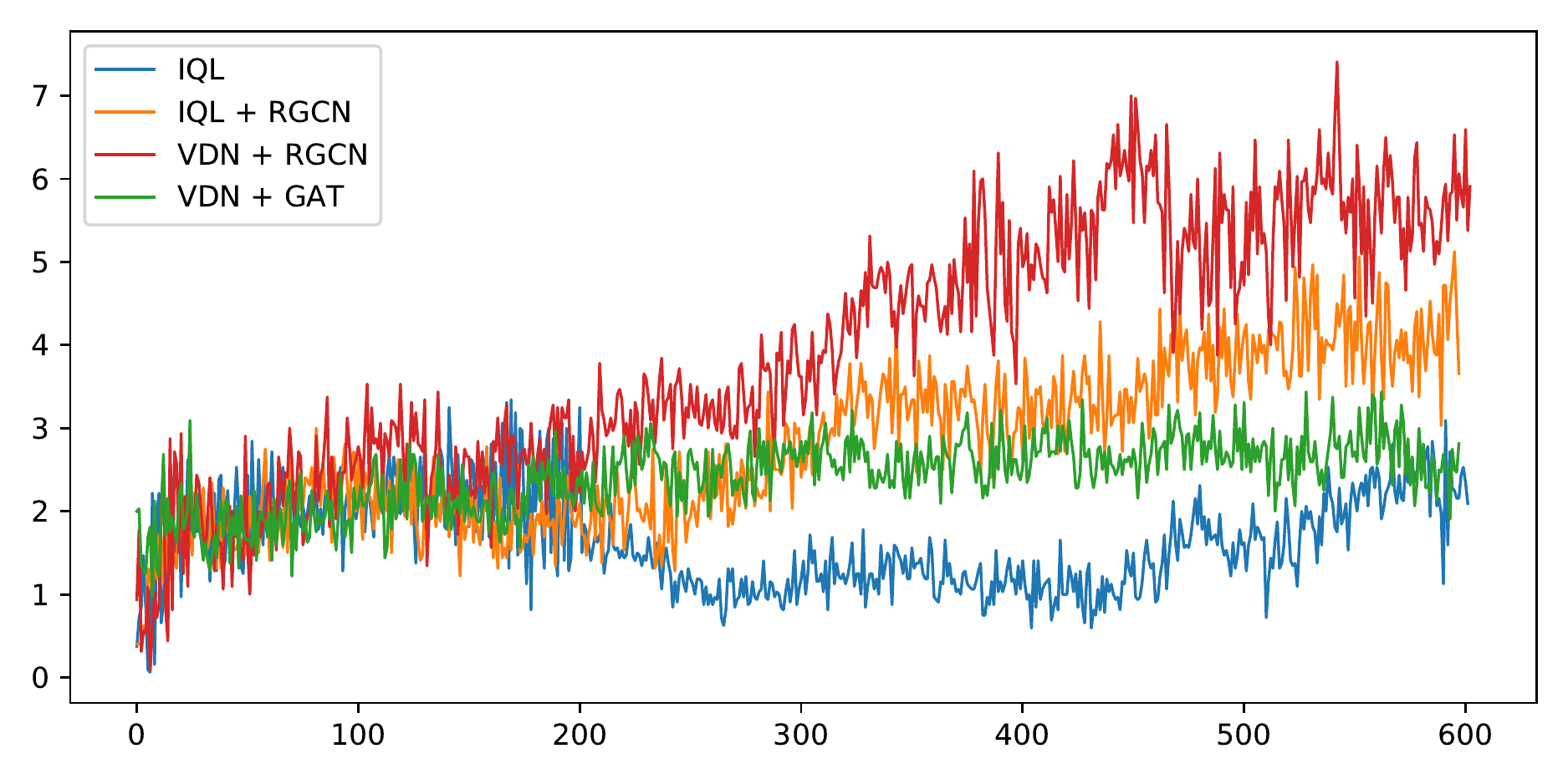}}
\end{figure}

The win rate and number of defeated enemies for the 6 million training steps in the 3s5z map are displayed in figure~\ref{fig:6M}. We can see that, in this scenario, agents only start to win episodes after approximately 3 million training steps (fig.~\ref{fig:wr_6M}). Also, only the models that implemented specialized communication demonstrated winning behavior, with the attention model with value decomposition winning an episode only after \~530 million steps.

In figure~\ref{fig:de_6M}, we see that the models with specialized communication achieve the highest performance, with the one employing value decomposition being the best of all. This suggests that the application of mixing strategies achieve an orthogonal increase in performance when applied with the proposed specialized agent communication method.

\section{Conclusion}

This work presented a method to model communication between heterogeneous agents in fully cooperative multi-agent deep reinforcement learning tasks through the use of relational graph convolutions. Agents are assigned to classes, in such a way that agents in the same class are homogeneous among themselves. The communication capabilities of the whole team in a given state is represented as a directed labeled agent graph, which encodes inter-agent class relations in the graph edges. Relational graph convolutions are used to specialize communication channels between pairs of agent classes using separate parameter matrices, which are reused every time two agents represented by the same ordered pair of classes communicates.

Experiments performed in a selection of scenarios from the StarCraft Multi-Agent Challenge show that the proposed method achieves equal or superior performance in all maps, when compared to models which use attention mechanisms for communication, or no communication at all. We also showed indications of how the proposed communication method can be combined with additive mixing, achieving even better results in the tested scenarios.

\section{Acknowledgments}

The authors acknowledge the S\~ao Paulo Research Foundation (FAPESP Grant 2019/07665-4) for supporting this project. This study was financed in part by the Coordenação de Aperfeiçoamento de Pessoal de Nível Superior - Brasil (CAPES) - Finance Code 001.

\bibliography{/home/dodo/Documents/Doutorado_bibtex.bib}

\end{document}